\pdfoutput=1
\documentclass[11pt]{article}

\usepackage{ACL2023}

\usepackage{times}
\usepackage{latexsym}

\usepackage[T1]{fontenc}
\usepackage[utf8]{inputenc}
\usepackage{graphicx}
\usepackage{microtype}

\usepackage{inconsolata}
\usepackage{caption}
\usepackage{subcaption}
\usepackage{enumitem}
\usepackage{ifthen}

\newboolean{anonymous}
\setboolean{anonymous}{true} % Set the boolean variable to true or false
\newcommand{\anonymousText}[2]{%
    \ifthenelse{\boolean{anonymous}}{#1}{\textit{#2}}%
}

\usepackage{multirow}
\usepackage{multicol}
\usepackage{amssymb}% http://ctan.org/pkg/amssymb
\usepackage{pifont}% http://ctan.org/pkg/pifont
\usepackage{booktabs}
\usepackage{graphics}
\usepackage{graphicx}
\usepackage{tablefootnote}
\usepackage{footnote}
\usepackage{soul}
\usepackage{fdsymbol}

\newcommand{\bl}[1]{\textcolor{black}{#1}}
\title{Forgotten Knowledge: Examining the Citational Amnesia in NLP}
  \author{
 {\hypersetup{linkcolor=black} Janvijay Singh\thanks{\ \ Equal contribution.}$^{*\clubsuit}$,   Mukund Rungta$^{*\clubsuit}$, %*}$, 
  Diyi Yang$^{\lozenge}$, Saif M. Mohammad$^{\Phi}$}\\
$^{\clubsuit}${Georgia Institute of Technology}, $^{\lozenge}${Stanford University},
$^{\Phi}${National Research Council Canada} \\
 \textcolor{darkblue}{\texttt{\{\href{mailto:iamjanvijay@gatech.edu}{iamjanvijay}, \href{mailto:mrungta8@gatech.edu}{mrungta8}\}@gatech.edu}},
  \texttt{\href{mailto:diyiy@stanford.edu}{diyiy@cs.stanford.edu}},\\
 \texttt{\href{mailto:saif.mohammad@nrc-cnrc.gc.ca}{saif.mohammad@nrc-cnrc.gc.ca}} \\
}

\begin{document}
\maketitle
\begin{abstract}
Citing papers is the primary method through which modern scientific writing
discusses and builds on past work.
Collectively, citing a diverse set of papers (in time and area of study) is an indicator
of how widely the community is reading.
Yet there is little work looking at broad temporal patterns of citation.
% e.g., the degree to
% which we cite recently published papers vs.\@ those published much earlier.
This work, systematically and empirically examines: \textit{How far back in time do we tend to go to cite papers?
How has that changed over time, and what factors correlate with this citational attention/amnesia?}
We chose NLP as our domain of interest, % as it is currently in a period of dramatic change; developments that are also likely impacting citation patterns. We 
and analyzed $\sim$71.5K papers % from the ACL Anthology 
to show and quantify several key trends in citation.
Notably, $\sim$62\% of cited papers are from the immediate five years prior to publication, whereas only $\sim$17\% are more than ten years old.
Furthermore, we show that the median age and age diversity of cited papers was steadily increasing from 1990 to 2014, but since then the trend has reversed, and 
\bl{current NLP papers have an all-time low temporal citation diversity.}
% dramatically so. 
Finally, we show that unlike the 1990s, the highly cited
papers in the last decade were also papers with the least citation diversity;
likely contributing to the intense (and arguably harmful) recency focus. 
% We now cite recent papers more profusely than ever before.
% Specifically, we found that Gini-index as computed over the age of referenced paper has increased by $\sim$8.57\% and $\sim$54.43\% years after 1998 and 2014. 
% We also show that median age of cited papers has dropped by $\sim$2 years since 2014, indicating a trend towards citing more recent papers in the NLP field.
% Interestingly, such a trend is more pronounced in journals compared to conferences and workshops.
% We believe our work 
% Our work provides a basis for broader self-reflection: e.g., how much we are seduced by the latest technological fads at the expense of ignoring important work from the more distant past, and what kind of work leaves a lasting long-term impact. 
\bl{ %To support these findings, we have made the 
Code, data, and a demo are available at the project homepage.
\footnote{Code, data: 
\anonymousText
{\url{https://github.com/iamjanvijay/CitationalAmnesia/}}
{The web link to the code and data is anonymized for review.\\}
}
\footnote{Online demo:  
\anonymousText
{\url{https://huggingface.co/spaces/mrungta8/CitationalAmnesia/}}
{The web link to the online demo is anonymized for review.}
}
}

\end{abstract}

\section{Introduction}
\label{sec:intro}

% -- role of citations; 
% -- importance of past paper citations
% -- time of change and rapid progress for NLP, DL, AI
% -- more important to look at past lessons: ethics; dont break things
% oTher context: digitalization, Google scholar, easy access; Googl paper conclusion
% -- how old and diversity

\textit{Study the past if you would define the future.}\\ 
\hspace*{5.4cm} --- Confucius\\
\noindent The goal of scientific research is to create a better future for humanity. 
To do this we innovate on ideas and knowledge from the past.
Thus, a central characteristic of the scientific method and modern scientific writing is to discuss other work: to build on ideas, to critique or reject earlier conclusions, 
to borrow ideas from other fields, and to situate the proposed work. %, etc. 
Even when proposing something that others might consider dramatically novel,
it is widely believed that these new ideas have been made possible because of a
number of older ideas \cite{verstak2014shoulders}.
\textit{Citation} (referring to another paper in a prescribed format) is the primary mechanism to point the reader to these prior pieces of work and also to assign credit for shaping current work \cite{Mohammad20c,rungta-etal-2022}. 
Thus, we argue that examining citation patterns across time
can lead to crucial insights into what we value, what we have forgotten, and 
what we should do in the future. %  how these have changed over time.
% what factors impact this citational attention/amnesia, what citation patterns are associated with different types of papers,} and  
% \textit{how citation patterns have changed over time}. 

% In this work, we systematically and empirically examine how far back we often cite papers, how such citation patterns change over time, and what factors affect our citation behaviors by analyzing temporal citation patterns of scientific papers. 

Of particular interest is the extent to which good older work is being forgotten --- \textit{citational amnesia}. More specifically, for this paper, we define citational amnesia as shown below:
\begin{quote}
\textit{Citational Amnesia}: the tendency to not cite enough relevant good work from the past (more than a few years old).    
\end{quote}
\noindent We cannot directly measure citational amnesia empirically because determining "enough", "relevance", and "good" require expert researcher judgment.
However, what we can measure is the collective tendency of a field to cite \textit{older} work. 
Such an empirical finding enables reflection on citational amnesia. A dramatic drop in our tendency to cite older work should give us cause to ponder whether we are putting enough effort to read older papers (and stand on the proverbial shoulders of giants). % (If we are not familiar with some work, then we are less likely to cite it.)
% Such conscious effort and a careful, systematic look at the related work will help us better reflect on whether the current rate of citing older papers is appropriate and how we as a field should move forward. 

Note that we are not saying that old work should be cited simply because it exists. We are saying that we should consciously reflect on the diversity of the papers we explore when conducting research. Diversity can take many forms, including reading relevant papers from diverse fields, by authors from diverse regions, and relevant papers published from various time periods --- the focus of this paper.   
Exploring a diverse set of papers allows us to benefit from important and diverse research perspectives. % However, if such a search leads to a low number of cite-worthy papers for a project, that is okay too.
Looking at older literature makes us privy to broader trends, and informs us in ways that are beneficial well beyond the immediate work.  

Historically, citational amnesia was impacted by various factors around access and invention. For example, the invention of the printing press in the year 1440 allowed a much larger number of people to access scientific writing \cite{eisenstein1985printing}. The era of the internet and digitization of scientific literature that began in the 1990s also greatly increased the ease with which one could access past work \cite{verstak2014shoulders}.  
However, other factors such as the birth of paradigm-changing technologies may also
impact citation patterns; ushering in a trend of citing very new work or citing work from previously ignored fields of work.
Such dramatic changes are largely seen as beneficial; however, strong tailwinds may also lead
to a myopic focus on recent papers and those from only some areas, at the expense of 
% good relevant work from other time periods and other areas 
benefiting from a wide array of work
\cite{pan2018memory,martin2016back}.

We choose as our domain of interest, papers on Natural Language Processing (NLP),
specifically those in the ACL Anthology.
This choice is motivated by the fact that NLP (and other related fields of Artificial Intelligence) are in a period of dramatic change: There are notable and frequent gains on benchmark  datasets; NLP technology is becoming increasingly ubiquitous in society; and new sub-fields of NLP such as Computational Social Science, Ethics and NLP, and Sustainable NLP are emerging at an accelerated rate.
The incredibly short research-to-production cycle and move-fast-and-break-things
attitude in NLP (and Machine Learning more broadly) has also led to considerable adverse outcomes for various sections of society, especially those with the least power \cite{buolamwini2018gender,article19_2021,mohammad2021ethics}.
% Thus, going beyond the most recent fads, and reflecting on the lessons and ideas from a broad spectrum of past research (from various fields of study) is arguably 
Thus reading and citing more broadly is especially important now.
% These developments are also likely impacting citation patterns.
% That said, we believe it is in the interest of all fields of study to reflect on the 
% questions addressed in this paper. %at the end of the first paragraph.

% While there is some work on examining citations from the perspective of 
% how much citations papers on different types of work and by different groups of people have received,
% there is no work looking at the temporal nature of citations.
% This work, for the first time, aims to identify broad temporal trends in citations by NLP papers. 
% and of NLP papers by NLP papers.

In this work, we compiled a temporal citation network of 71.5K NLP papers % from the ACl Anthology 
that were published between 1990 and 2021, along with their meta-information such as the number of citations they received in each of the years since they were published --- 
% We will refer to this dataset as 
the \textit{Age of Citations (AoC) dataset}. 
We use AoC to answer a series of specific research questions on \textit{what we value, what we have forgotten, what factors are associated with this citational attention/amnesia, what are the citation patterns of different types of papers,} and  
\textit{how these citation patterns have changed over time}. 
Finally, we show that many of the highly cited papers from the past decade have very low
temporal citation diversity; and because of their wide reach, may have contributed to the intense
recency focus in NLP.
%\footnote{Note that this work will not attempt to automatically identify individual pieces of good work that ought to be cited more. That requires the judgment of an expert in the field and different stakeholders may find different pieces of work important.}
% temporal citation patterns: How far back in time we go to cite papers? What is the distribution of citations a paper is expected to get in the years after it is published? 
% The answers to these questions not only shed light on various temporal citation patterns in NLP, but they also
% enable further inquiry and future work into bigger questions such as what kind of work leaves a lasting long-term impact. 
All of the data and code associated with the project will be made freely available on the project homepage.

\section{Related Work}

In the broad area of Scientometrics (study of quantitative aspects of scientific literature),
citations and their networks have been studied from several perspectives, including:
paper quality \citep{buela2010analysis}, field of study \citep{costas2009scaling}, novelty, length of paper \citep{antoniou2015bibliometric, falagas2013impact},  impact factor \citep{callaham2002journal}, 
venue of publication \citep{callaham2002journal,WahleRMG22},
language of publication \citep{lira2013influence},  and  number of authors  \citep{della2008multi, bosquet2013academics}, collaboration \citep{nomaler2013more}, self-citation \citep{costas2010self}, as well as author’s reputation \citep{collet2014does}, affiliation \citep{sin2011international,lou2015does}, 
geographic location \cite{nielsen2021global,lee2010author,pasterkamp2007citation,paris1998region}, %rungta-etal-2022},
% paris1998region, nishioka2022does,rungta-etal-2022},
gender, race and age \citep{ayres2000determinants,leimu2005determines,chatterjee2021gender, llorens2021gender}.

However, there has been relatively little work exploring the temporal patterns of citation.
% -- Google Scholar Giants paper\\
% -- Citation decay paper\\
\citet{verstak2014shoulders} analyzed scholarly articles published in 1990--2013 to show that the percentage of older papers being cited steadily increased from 1990 to 2013, for seven of the nine fields of study explored.
(They treated papers that were published more than ten years before a particular citation as \textit{old papers}.) 
For Computer Science papers published in 2013, on average, 28\% of the cited papers were published more than ten years before. This represented an increase of 39\% from 1990. 
They attributed this increasing trend in citing old papers to the ease of access of scientific literature on the world wide web, as well as the then relatively new scientific-literature-aggregating services such as Google Scholar.

\citet{PAROLO2015734} analyzed about 25 million papers from Clinical Medicine, Molecular Biology, Physics, and Chemistry
published until 2014 
to show that typically the number of citations a paper receives per year increases in the years after publication, reaches a peak, and then decays exponentially. Interestingly they showed that this rate of decay was increasing in the more recent papers of their study. They attribute this quicker decay (or more ``forgetting'' of recent papers) to the substantial increase in the number of publications; a lot more papers are being published, and due to the limited attention span of subsequent researchers, on average, papers are being forgotten faster.  

Past work on NLP papers and their citations includes work on gender bias \cite{schluter2018glass,vogel-jurafsky-2012-said,mohammad-2020-gender},
 author location diversity \cite{rungta-etal-2022}, 
 author institution diversity \cite{abdalla2023elephant},
 and 
on broad general trends such as average number of citations over time and by type of paper \cite{Mohammad20c,Mohammad20b,WahleRMG22}. 

\citet{bollmann-elliott-2020-forgetting} were the first to explore the recency bias of citations in NLP papers. They % analyzed papers published between 2010 and 2019 in the ACL Anthology to 
showed that the ACL Anthology papers published between 2017 and 2019 cited more recent work than papers published between 2010 and 2014. Question 3 in Section 4 of our paper is of the same spirit that was explored in their work; however, our work examines a  much larger spread of NLP papers (published between 1965 and 2021). This will shed light on the reproducibility of those findings and, more importantly, determine the broader trajectory of temporal citation patterns (from the start of ACL to present day). 
% which we validated and obtained consistent findings via our Research Question 3 in Section 4 using a much larger spread of NLP papers from 1965 to 2021 over 5 decades. 
% More importantly, 
Additionally, our work introduces a new citation age diversity metric to quantify the degree of spread of citations over time, as well as an interactive online demo system to visualize the citation age diversity of any paper. Going beyond how overall citation patterns have changed over time, our work takes a deep dive into six other novel research
questions, notably  around temporal citation patterns in sub-areas of NLP, of cited topics, and across sparsely and highly cited papers.
% Question 3 in Section 4 of our paper is of the same spirit that was explored in their work; however, our work examines a  much larger spread of NLP papers (published between 1965 and 2021) and introduces a new citation age diversity metric that better captures the degree of spread of citations over time. 
% We also explore six other questions pertaining to citation patterns  (notably in sub-areas of NLP and across low- and high-citation papers).

% Further, it is unclear whether even the broader trends in the Sciences until about 2014 (discovered by \citet{verstak2014shoulders} and \citet{PAROLO2015734}), still hold true. 
% discussed in the Introduction), 

\section{Dataset}
The ACL Anthology (AA) Citation Corpus \cite{rungta-etal-2022} % as a starting point for our dataset curation process. 
% This corpus, released by \citet{rungta-etal-2022}, 
contains meta data (paper title, year of publication, and venue, etc.\@ %of publication 
for the 71,568 papers in the ACL Anthology repository (published until January 2022). 
% However, the corpus does not include information about the papers cited in these papers. 
% To address this gap, 
We used the Semantic Scholar API\footnote{ {\url{https://www.semanticscholar.org/}}} to gather the references for each paper in the AA Citation Corpus, using the paper's unique Semantic Scholar ID (SSID).
This allowed us to obtain additional information about the \textit{cited papers}, such as their title, year of publication, and venue of publication. 
Note that these cited papers may or may not be part of AA.
To study the dynamics of citations over time, we constructed year-wise citation networks using the data collected.
Specifically, we % augmented 
created the citation networks for every year from 1965 to 2001. 
% a given year by including papers published in that year, as well as the citing edges to papers in the citation network from the previous year. 
This representation of citation data allows us to answer several interesting questions, such as the number of citations a paper receives in a particular year after its publication.
We refer to this dataset as \textit{Age of Citations (AoC) dataset}.
% which we will release for the community to perform various analyses that rely on the temporal nature of citations.

% As of January 2022, the ACL Anthology (AA) had 71,568 papers.\footnote{{\url{https://aclanthology.org/}}} We used the AA Citation Corpus released by X. This dataset contains information about the paper title, names of authors, year of publication, and venue of publication for each of these papers from the repository. This dataset does not contain information about the cited papers. Therefore, we augmented this dataset further to include the references for each paper. This information is gathered using the Semantic Scholar API.\footnote{ {\url{https://www.semanticscholar.org/}}}
% Not all cited papers are present in the AA Citation Corpus, and thereby we do not have meta-information about these papers. We further use the unique Semantic Scholar ID (SSID) of each paper to extract information like paper title, year of publication, and venue of publication. In this way, we were able to gather all the information about the cited papers which are required for our next steps.

% Mention something about how the remainder of the analysis is done on papers from 1990 onwards because the number of NLP papers published before that is quite small.

\section{Age of Citation}
% \diyi{i found the current way of writing taking too much space. basically for each RQ, you talk about the exp setup, but they share a lot of things in common. Can we put all methods together and then the results only focus on results?}
% Saif: I think the reorging that J and I discussed yesterday fixes most of that. 
% Also, putting more about creating the dataset in Section 3 should help with this.
% We can see then whetehr a separate "Methods" section is needed. I am not sure it is.

We used the \textit{AoC dataset} to answer a series of questions on how research papers are cited and the trends across years. 
% We start with a look at the average number of papers cited by a AA paper, followed by analyzing the distribution and possible factors impacting it.

% ----------------R1-START---------------
% \begin{figure}[h]
%     \centering
%     % \includegraphics[width=0.9\textwidth]{LaTeX/venue_country_dst.png}
%     \includegraphics[width=1\columnwidth]{figures/r4_reference_distribution_overall.png}
%     \caption{R1: Average number of papers referenced by an AA paper = 20.6. Plots showing the trend of the average number of papers referenced by an AA paper across different years.}
%     \label{fig:r1_new_count_ref}
% \end{figure}
% \begin{figure}[h]
%     \centering
%     % \includegraphics[width=0.9\textwidth]{LaTeX/venue_country_dst.png}
%     \includegraphics[width=1\columnwidth]{figures/r4_reference_distribution_venues.png}
%     \caption{R1: Plots showing the trend of the average number of papers referenced by an AA paper across different years for different venues of publication.}
%     \label{fig:r1_new_count_ref_venue}
% \end{figure}
\begin{figure}[t]
    \centering
    \includegraphics[width=1\columnwidth]{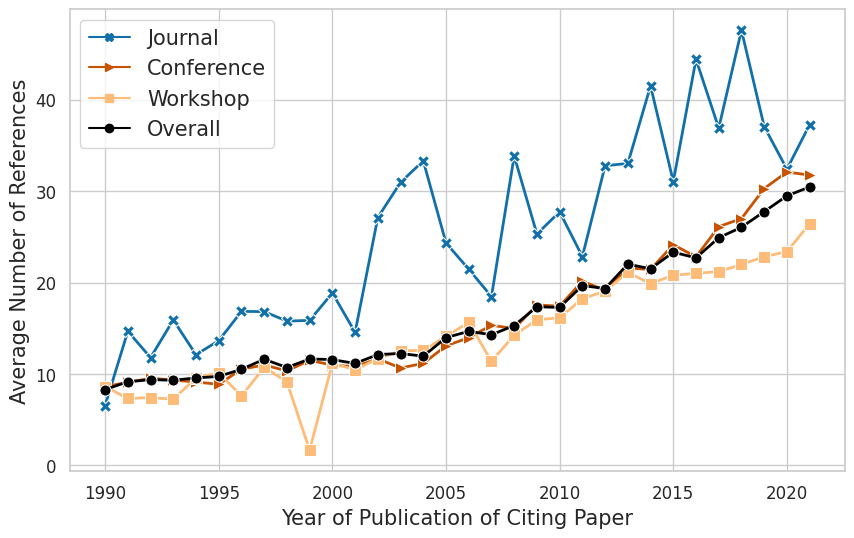}
    \caption{Average number of unique references in an AA paper published in different years.}
    %    \textbf{Q1}: 
    % Trend of average number of papers referenced by AA paper across different years, for overall and grouped by different types of publication.}
    \label{fig:r1_new_count_ref_combined}
\end{figure}
\vspace{2mm}
\noindent
\textbf{Q1. What is the average number of unique references in the AA papers? How does this number vary by publication type, such as workshop, conference, and journal? Has this average stayed roughly the same or has it changed markedly over the years?}\\[-10pt]

% Refer to the following figure in the text: 
% \ref{fig:r1_new_count_ref} 
% \ref{fig:r1_new_count_ref_venue}
% \ref{fig:r1_new_count_ref_combined}
\noindent
\textbf{Ans.} 
We % begin our analysis by 
calculated the average number of unique references for all papers in the \textit{AoC dataset}, as well as %the average number of citations grouped by 
for each publication type (workshops, conferences, and journals). We then binned all papers by publication year, computed the mean and median for each bin %number of citations for each bucket, and calculated the averages 
for each year. 

\begin{table}[t]
% \small
\begin{center}
\setlength{\tabcolsep}{5pt}%Tighter
% \scalebox{0.95}{
\scalebox{0.8}{
\begin{tabular}{lcc}
\toprule
& \textbf{Mean}
& \textbf{Median} \\
% & \textbf{Mode} \\
\midrule
\textit{Journal} &	 23.24  & 15   \\
\textit{Conference} &	21.11  & 19   \\
\textit{Workshop} &	 19.07  & 17  \\
\textit{Overall} &	20.63  & 18   \\
\bottomrule 
\end{tabular}
}
\end{center}
\vspace*{-3mm}
\caption{\label{tab:avgnumcites}
Mean and median of the number of unique references in an AA paper.}
% in AA, also broken down by publication types.}
\vspace*{-3mm}
\end{table}

\paragraph{Results} 
% We found that, on average, an AA paper refers to (cites) 20.6 papers. (The median is 18.) % and the mode  1.
The scores % for the full dataset and for different paper types 
are shown in Table \ref{tab:avgnumcites}. 
Figure \ref{fig:r1_new_count_ref_combined} shows how the mean has changed across the years.\footnote{The numbers of AA papers published each year until 1990 were rather low, and so in Figure \ref{fig:r1_new_count_ref_combined}, we only show the trajectory from 1990. However, note that the numbers generally increase even from 1965 to 1990.}
% although there were big fluctuations from year to year.} 
% The black colored line indicates the trend of all the papers published in the concerned year. Similarly, blue, red, and orange colored lines are for journal, conference, and workshop respectively. 
% during that period which makes the average number of references highly inconsistent. 
% However, after 1990, a considerable number of papers were published in the AA and the graph post-1990 follows a more consistent upward trend.}
% Figure \ref{fig:r1_new_count_ref_combined} shows the trend of average number of papers referenced by an AA paper with respect to their publication year. 
The graph shows a general upward trend. The trend seems roughly linear until the mid 2000s, at which point we see that the slope of the trend line increases markedly. 
Even just considering the last 7 years, there has been a 41.74\% increase in referenced papers in 2021 compared to 2014. 
% Additionally, it can be observed that the number of references per paper has continually increased over time. 
% Figure \ref{fig:r1_new_count_ref_combined} also shows the trend of average number of papers referenced by an AA paper for each of the venues.

Similar overall trends can be observed when papers are grouped by publication type.
Not surprisingly, the longer journal articles cite markedly more papers than conference and workshop papers.
The plot for conferences and workshops is relatively smooth compared to journal articles. 
This is because the number of papers for each year in journals is far less. 
For example, in the year 2015, only 139 papers were published in journals, whereas 1709 and 983 papers were published in conferences and workshops respectively.\\[-16pt]

\paragraph{Discussion} 
The steady increase in the number of unique references from 1965 % onwards, 
is likely because of the increasing number of relevant papers as the field develops and grows. However, it is interesting that this growth has not plateaued even after 55 years.
% The continual increase of average number of references can be attributed to factors such as page limit, the availability of more relevant work in the field, and others.
\bl{By the late}-2000s, with the advent of widely accessible electronic proceedings, *ACL venues started experimenting with more generous page limits:
relaxing it from a strict 8 pages to first allowing one or two additional pages for references to eventually allowing unlimited pages for references.\footnote{\bl{In 2008, EMNLP became the first major NLP venue to allow an extra page for references. (ACL followed in 2009.)}}
Other %changes 
\bl{factors} that may have contributed to more papers being referred to (cited) within a paper, include: an additional page for incorporating reviewer comments,  allowing Appendices, \bl{and the inclusion of an increasing number of experiments}.
% This is probably because the field of NLP has experienced a significant increase in the number of referenced papers in the last several years, particularly due to the advent of deep learning. 
% The plots for journals, conferences, and workshops follow a roughly similar trend, but 

% ----------------R1-END---------------

% ----------------R2-START---------------
% \begin{figure}[h!]
%     \centering
%     % \includegraphics[width=0.9\textwidth]{LaTeX/venue_country_dst.png}
%     \includegraphics[width=1\columnwidth]{figures/r1_aoc_distribution_overall.png}
%     \caption{R2: Overall Distribution of AoC for all papers in AA. For a given age in the x-axis, the corresponding value in the y-axis represents the percentage of total references having that age.}
%     \label{fig:r2_new_aoc_dist}
% \end{figure}
% \begin{figure}[h]
%     \centering
%     % \includegraphics[width=0.9\textwidth]{LaTeX/venue_country_dst.png}
%     \includegraphics[width=1\columnwidth]{figures/r1_aoc_distribution_venues.png}
%     \caption{R2: Distribution of AoC for all papers in AA based on venue of publication. For a given age in the x-axis, the corresponding value in the y-axis represents the percentage of total references from the venue having that age.}
%     \label{fig:r2_new_aoc_distribution_venue}
% \end{figure}
\vspace{2mm}
\noindent
\textbf{Q2. On average, how far back in time do we go to cite papers? As in, what is the average age of cited papers? What is the distribution of this age across all citations? How do these vary by publication type?}\\[-10pt]

% What is the average AoC of papers? By JCW? (Put these in a table: 12 number 4 by 3) What are their distributions?

% figures to be reffered here:
% \ref{fig:r2_new_aoc_distribution_combined}
% \ref{fig:r2_new_aoc_dist}
% \ref{fig:r2_new_aoc_distribution_venue}
\noindent
\textbf{Ans.} %In order to gain insights into the distribution of citations in the \textit{AoC dataset}, we examined the age of citations associated with all papers present in the dataset.
If a paper $x$ cites a paper $y_i$, then the age of the citation (\textit{AoC}) 
% of paper $y_i$, within a citing paper $x$ 
is taken to be the difference between the  year of publication (\textit{YoP}) of $x$ and % YoP of 
$y_i$: 
% Specifically, this age $AoC (x,y_i)$ can be expressed in terms of year of publication, $YoP$, as follows:
\begin{equation}
\textrm{\it AoC} (x,y_i) = \textrm{\it YoP} (x) - \textrm{\it YoP} (y_i)
\end{equation}
\noindent We calculated the AoC for each of the citations in the AoC dataset.
  For each paper, we also calculated the mean $AoC$ %($mAoC(x)$ for short) 
  of all papers cited by it:\\[-10pt] 
 % Specifically, we define $mAoC$ as:
\begin{equation}
mAoC(x) = \frac{1}{N} \sum_{i=1}^{N} AoC (x,y_i)
\end{equation}
%\noindent here $y_1, y_2, y_3 \ldots, y_N$ refer to the $N$ papers cited by paper $x$. 
\noindent here $N$ refers to the number of papers cited by $x$.

\begin{figure}[t]
    \centering
    \includegraphics[width=1\columnwidth]{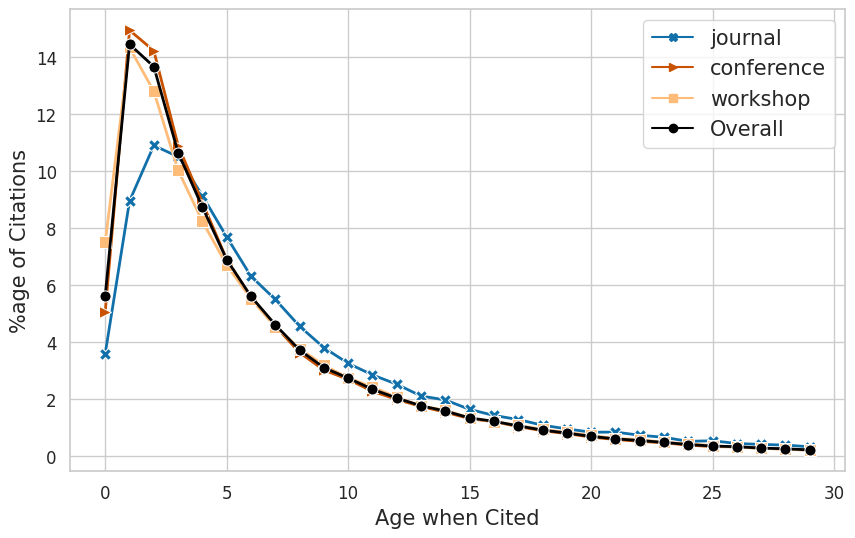}
    \caption{
    % \textbf{Q2}: 
    % Overall and publication-type-wise 
    Distribution of $AoC$ for papers in AA (overall and by publication type).}
    % The x-axis represents the $AoC$, while the y-axis represents the percentage of total citations with that $AoC$.}
    \label{fig:r2_new_aoc_distribution_combined}
 %   \vspace*{-3mm}
\end{figure}

\paragraph{Results} The average mAoC for all the papers in the \textit{AoC dataset} is 6.01. % 7.02.
% age of citations or $AoC$ for all papers in the \textit{AoC dataset} is W. 
% When considering different publication venues, such as workshops, conferences, and journals, the average AoC is X, Y, and Z respectively. 
% The scores were 8.16 for journal articles, 6.93 for conference papers, and 7.01 for workshop papers.
The scores were 7.16 for journal articles, 5.91 for conference papers, and 6.01 for workshop papers.
Figure \ref{fig:r2_new_aoc_distribution_combined} shows the distribution of $AoC$s in the dataset
across the years after the publication of the \textit{cited} paper (overall, and across publication types).
For example, the y-axis point for year 0 corresponds to the average of the percentage of citations papers received 
in the same year as it they were published. The y-axis point for year 1 corresponds to the average of percentage of citations the papers received 
in the year after they were published. And so on. 
% The figure provides a visual representation of the distribution of citations (as percentages of total citations) across the age of cited papers.
% Upon examination of the figure, several key observations can be made. 
% Firstly, it can be seen that 

Observe that the majority of the citations are 
for papers published one year prior,
% cited in the year immediately after publication
(AoC = 1). 
% This suggests that the majority of citations in the dataset are made to papers that have been recently published. 
This is true for conference and workshop subsets as well, but %when looking at citations by 
in journal papers, the most frequent citations are for papers published two years prior.
Overall though all the arcs have a similar shape, rising sharply from the number in year 0 to the peak value and then dropping off at an exponential rate in the years after the peak is reached.
For the full set of citations, this exponential decay from the peak has a half life of about 4 years.
Roughly speaking, the line plot for journals is shifted to the right by a year compared to the line plots for conferences and workshops. It also has a lower peak value and its citations for the years after the peak are at a higher percentage than those for  conferences and workshops. 
Additionally, citations in workshop papers  have the highest percentage of current year citations (age 0), whereas citations in journal article have the lowest percentage of current year citations. 
% number of papers with an age of 0, indicating that workshops and conferences tend to cite more recent papers compared to journals.

\begin{figure}[t]
    \centering
    \includegraphics[width=1\columnwidth]{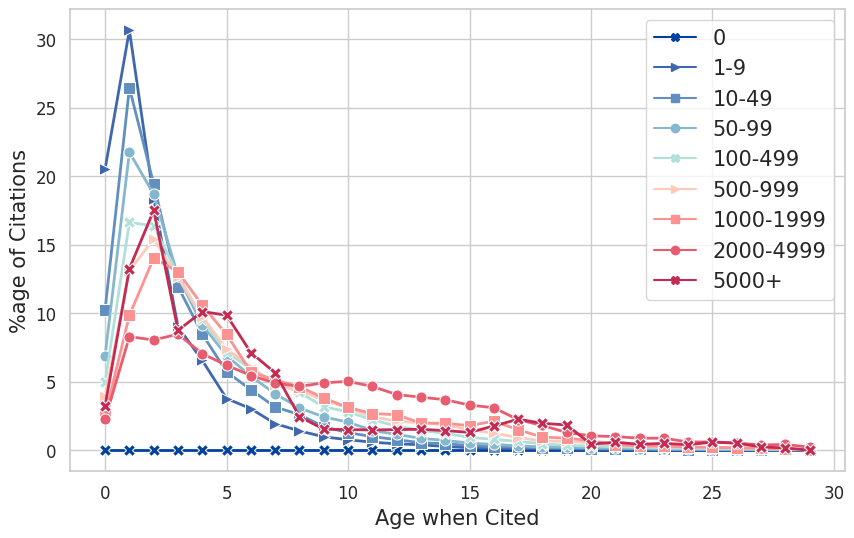}
    \caption{
    % \textbf{Q2}: 
    % Overall and publication-type-wise 
    \bl{Distribution of $AoC$ for AA papers with different citation counts (shown in legend).}}
    % The x-axis represents the $AoC$, while the y-axis represents the percentage of total citations with that $AoC$.}
    \label{fig:r2_new_aoc_distribution_bins}
 %   \vspace*{-3mm}
\end{figure}

\bl{Analogous to Figure \ref{fig:r2_new_aoc_distribution_combined}, Figure \ref{fig:r2_new_aoc_distribution_bins} presents the distribution of AoCs, albeit broken down by the total citations received by a paper. It is worth noting that the distribution leans more towards the right for papers with a higher number of citations. This shows that papers with a higher citation count continue to receive significant citations even far ahead in the future, which is intuitive.}

\paragraph{Discussion} Overall, we observe that papers are cited most in years immediately after publication, and their chances of citation fall exponentially after that. The slight right-shift for the journal article citations is likely, at least in part, because journal submissions have a long turn-around time from the first submission to the date of publication (usually between 6 and 18 months). \bl{A list of the oldest papers cited by AA papers is available on the project's GitHub repository.} % \footnote{\url{https://github.com/iamjanvijay/CitationalAmnesia/blob/main/dataset/cited_papers.tsv}}}

% Secondly, it can be observed that cited papers with an age of 0 are greater in number than cited papers with an age greater than 5. This implies that papers that have just been published tend to receive more citations in comparison to older papers.
% Additionally, the age of cited papers roughly follows an exponential decay, with a half-life of nearly 4 years. 
%  which might be due to longer review period of journal papers.

% Figure \ref{fig:r2_new_aoc_distribution_combined} also illustrates the distribution of $AoC$ for papers published in different venues such as workshops, conferences and journals. 
% The majority of papers have an age of 1-2 years, indicating that papers published recently are cited more often. 

% This analysis has revealed intriguing patterns in the age of cited papers, at aggregate. 
% The findings from this analysis motivated us to further investigate the dynamics of citation patterns by delving deeper into the evolution of age of citations over time, which we examine in our next research question. 
% A deeper understanding of the evolution of $AoC$ can provide valuable insights into how the popularity of papers changes over time and provide a deeper understanding of the dynamics of citation patterns in the field of NLP.

% ----------------R2-END---------------

% ----------------R3-START--------------
% \begin{figure}[h]
%     \centering
%     % \includegraphics[width=0.9\textwidth]{LaTeX/venue_country_dst.png}
%     \includegraphics[width=1\columnwidth]{figures/r2_gini.png}
%     \caption{R2: Gini coefficient trend for the overall distribution of $mAoC$ across years.}
%     \label{fig:r3_new_gini}
% \end{figure}
% % venue wise figures
% \begin{figure}[h]
%     \centering
%     % \includegraphics[width=0.9\textwidth]{LaTeX/venue_country_dst.png}
%     \includegraphics[width=1\columnwidth]{figures/r6_gini_venue.png}
%     \caption{R2: Plot of Gini coefficient for different venues across time. }
%     \label{fig:r3_new_gini_venue}
% \end{figure}
\begin{figure}[t]
    \centering
    \includegraphics[width=1\columnwidth]
    % {figure_final/r3_gini_trend.png}
    {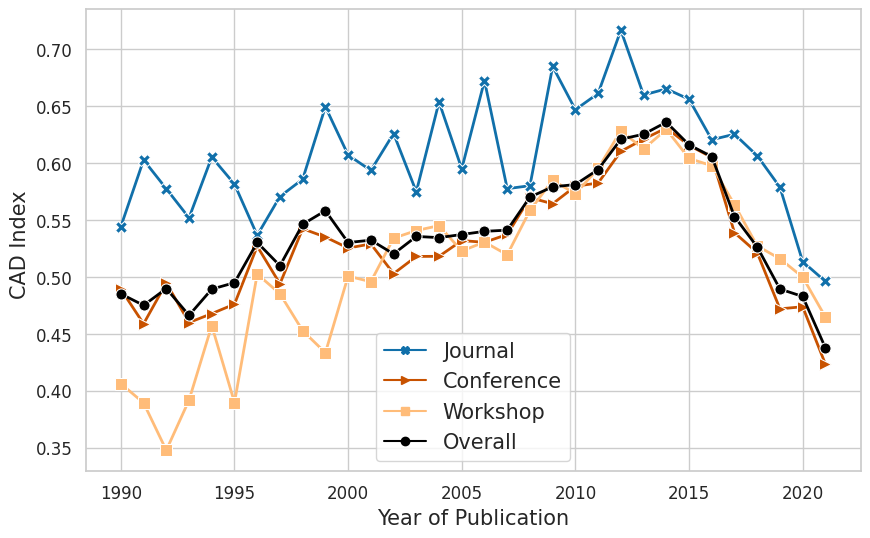}
    \caption{ %\textbf{Q3}: 
    Citation Age Diversity Index
    % Gini coefficient of mAoC  
     %for the overall and publication-type-wise distribution of $AoC$ 
    across years. }
    \label{fig:r3_new_gini_combined}
   \vspace*{-3mm}
\end{figure}
\vspace{2mm}
\noindent
\textbf{Q3. What is the trend in the variation of $\pmb{AoC}$ over time and how does this variation differ across different publication venues in NLP?}\\[-10pt]

\noindent
\textbf{Ans.} 
To answer this question, we split the papers into bins corresponding to the year of publication, and then examined the distribution of $mAoC$ in each bin.
\bl{
We define a new metric called the \textit{Citation Age Diversity (CAD) Index}, which measures the diversity in the $mAoC$ for a set of papers. 
In simpler terms, a higher \textit{CAD Index} indicates that $mAoCs$ covers a broader range, implying that the cited papers span a wider time period of publication.
This metric offers valuable insights into the temporal spread of scholarly influence and the long-term impact of research.
Precisely, the \textit{CAD Index} for a bin of papers $b$, is defined using the Gini Coefficient as follows:
\begin{equation}
CAD(b) = 1 - \sum_{i=1}^{N} \sum_{j=1}^{N}\frac{\left|mAoC(b_i)-mAoC(b_j)\right|}{2 N^2 \bar{b}}
\end{equation}
here, $b_i$ corresponds to $i^{th}$ paper within bin $b$, $N$ denotes the total number of papers in bin $b$ and $\bar{b}$ represents the mean of $mAoC$ of papers' associated with bin $b$. 
A \textit{CAD Index} close to 0 indicates minimum temporal diversity in citations (citing papers from just one year), whereas a \textit{CAD Index} of 1 indicates maximum temporal diversity in citations (citing papers uniformly from past years).
In addition to \textit{CAD Index}, we also compute median $mAoC$ of each such yearly bin.
The results for both \textit{CAD Index} and median $mAoC$ have roughly identical trends across the years. We discuss the \textit{CAD Index} analysis below. (The discussion of the median $mAoC$ results is in the Appendix \ref{sec:supplementary_1}.)}

\paragraph{Results}
% Figure \ref{fig:r3_new_gini_combined} shows the diversity of $mAoC$ scores across years (higher Gini coefficient indicates low diversity), and across different publication types. 
Figure \ref{fig:r3_new_gini_combined} shows the \textit{CAD Index} across years (higher \textit{CAD Index} indicates high diversity), and across different publication types. 
% The plot also shows Gini coefficients for each publication type.
The \textit{CAD Index} plot of Figure \ref{fig:r3_new_gini_combined} shows that the temporal diversity of citations had an increasing trend from 1990 to 2014, but the period from 1998 to 2004, and 2014 to 2021 (dramatically so) were periods of decline in temporal diversity (decreasing \textit{CAD Index} scores). 
% In addition to studying the recency of citations, we also investigate the diversity of citations using Figure \ref{fig:r3_new_gini_combined}. 
% Our analysis of the gini coefficient reveals an upward trend from 2014 to 2021 as well as from 1999 to 2002. 
These intervals coincide with the year intervals in which we observed a decreasing trend in median $mAoC$ of published papers (discussed in the Appendix).
% This suggests that the diversification in $mAoC$ has been on a decreasing trend in the year intervals 1999-2002 and 2014-2021.
% In other words, papers published in 
This suggests that the increase or decrease in diversity is largely because of the decreased or increased focus on papers from recent years, respectively.
% papers from these years tend to cite papers from a narrower range of publication years compared to papers published in other years. Similarly, from the plot \ref{fig:r3_new_gini_combined}, we can see two segments for all the venues. 

The \textit{CAD Index} plots by publication type all have similar trends, with journal paper submissions consistently having markedly \bl{higher} scores (indicating markedly higher temporal diversity)
across the years studied. However, they also seem to be most impacted by the trend since 2014 to cite very recent papers. (\textit{CAD Index} not only goes back to the 1990 level, but also undershoots beyond it.)
% We observe that from 1990 to 2014, the Gini coefficient dropped almost linearly for all venues. 
% For conferences, the average Gini coefficient dropped from 0.5108 to 0.3693 over this period. 
% This decrease in the Gini coefficient suggests diversification in research, as papers were being cited from all years roughly comparably. 
% However, in the second segment from 2014 to 2021, the Gini coefficient of diversification increased almost linearly for all venues. 
% For conferences, the average Gini coefficient increased from 0.3693 to 0.5766 from 2014 to 2021. 
% This steep increase in the trend showcases a myopic view of the age of citation distribution where most references belong to a certain time period. 

\paragraph{Discussion} Overall, we find that all the gains in temporal diversity of citations from 1990 to 2014 (a period of 35 years), have been negated in the 7 years from 2014. 
This change is driven largely by the deep neural revolution in the early 2010's and strengthened further by the substantial impact of transformers on NLP and Machine Learning.
Interestingly, our results until 2013 are in line with what \citet{verstak2014shoulders} found for many fields of study, but since 2014 there has been a marked shift in trends in NLP. We hope future work will explore whether similar shifts in trends have occurred in other fields. Our results add to (and are consistent with) the mean-citation age results found by \citet{bollmann-elliott-2020-forgetting}, who examined mean citation age between 2010 and 2019. Our analysis of the broader period (from 1965 to 2021), situates those results in the overall trajectory of how temporal citation patterns have evolved since the beginning of the Association of Computational Linguistics to the present period. Additionally, the new CAD Index metric quantifies the degree temporal citation diversity as opposed to the recency focus of citations captured by mean citation age.

% for papers published between 2017 and 2019.
% The experience of various NLP researchers suggests that this change is driven by the advent of the deep neural revolution in the early 2010's and the strengthened .

% We noted that this trend could be attributed to the recent surge in popularity of neural networks, which has led to a polarization towards citing only recent papers and specifically those involving neural methods. 
% Additionally, it is worth noting that the first segment spanned across 25 years, whereas the second segment achieved a similar Gini coefficient in just 8 years.
% Overall, these observed variations in $mAoC$ raises the question of identifying the factors responsible for it. In our next research questions, we aim to investigate the potential factors that may contribute to trends in $mAoC$.

% ----------------R3-END---------------

% ----------------R4-START---------------
\begin{figure}[t]
    \centering
    \includegraphics[width=1\columnwidth]{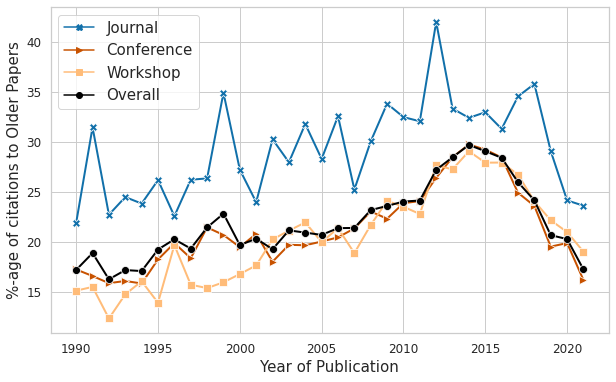}
    \caption{ %\textbf{Q4}: 
    Percentage of citations in AA papers where the cited paper is at least 10 years old.}
    %overall and broken down by publication type.}
    \label{fig:r4_new_old_refs}
\end{figure}
\vspace*{-5mm}
\vspace*{6mm}
\noindent
\textbf{Q4. What percentage of cited papers are old papers? How has this varied across years and publication venues? }\\[-10pt]
% What percentage of papers cited are old papers (papers > 10 years)? What are the scores for JCW? (We can show a table for this: rows are J, C, W, All; columns are 1990, 2014, 2021)

\noindent
\textbf{Ans.}
% To study the trend of citing \textit{older papers} in NLP, similar to 
Just as \citet{verstak2014shoulders}, we %began by defining 
define a cited paper as \textit{older} if it was published at least ten years prior to the citing paper.
We then divided all AA papers into groups based on the year in which they were published. 
For each AA paper, we determined the number of citations % that were attributed to older 
to older papers. 
% Using this information, we calculated the percentage of total citations for each year that were attributed to older papers. 
% Additionally, we also examine this trend across various venue types. This analysis is presented in Figure \ref{fig:r4_new_old_refs}.

\paragraph{Results} 
Figure \ref{fig:r4_new_old_refs} shows the percentage of older papers cited by papers published in different years.
% reveals several interesting patterns in terms of citing of \textit{older papers} in the field of NLP.
% Across all AA papers, the percentage of citations to older papers 
Observe that this percentage increased steadily from 1990 to 1999, before decreasing until 2002. 
After 2002, the trend of citing older papers picked up again; reaching an all time high of $\sim$30\% by 2014. 
However, since 2014, the percentage of citations to older papers has dropped dramatically, falling by 12.5\% and reaching a historical low of $\sim$17.5\% in 2021. 
% When examining this trend across different venue types, such as conferences and workshops, a similar pattern is observed.
Similar patterns are observed for different publication types.
However, we note that a greater (usually around 5\% more)  percentage of a journal paper's citations are to older papers, than in conference and workshop papers.
% for journal publications, the trend appears more irregular and jagged, which can be attributed to the limited number of papers published in journals.

\paragraph{Discussion} These results confirm that the trends in diversity discussed in Q2 are aligned with the trends in citing older papers.
% Similar to the trend of temporal diversity of citations, we observed a significant increase in the citing of older papers from 1990 to 2014, reaching an all-time high. 
% However, in the past 7 years, this trend has been reversed, with the fraction of old papers being cited reaching an all-time low. 
This dramatic drop in citing older papers since 2014 can largely be attributed to the explosion of paper count and the paradigm shift in the field of NLP brought on by deep learning and transformers.

% ----------------R4-END---------------

% ----------------R5-START--------------
\begin{figure*}[!ht]
    \centering
    \includegraphics[width=2\columnwidth]{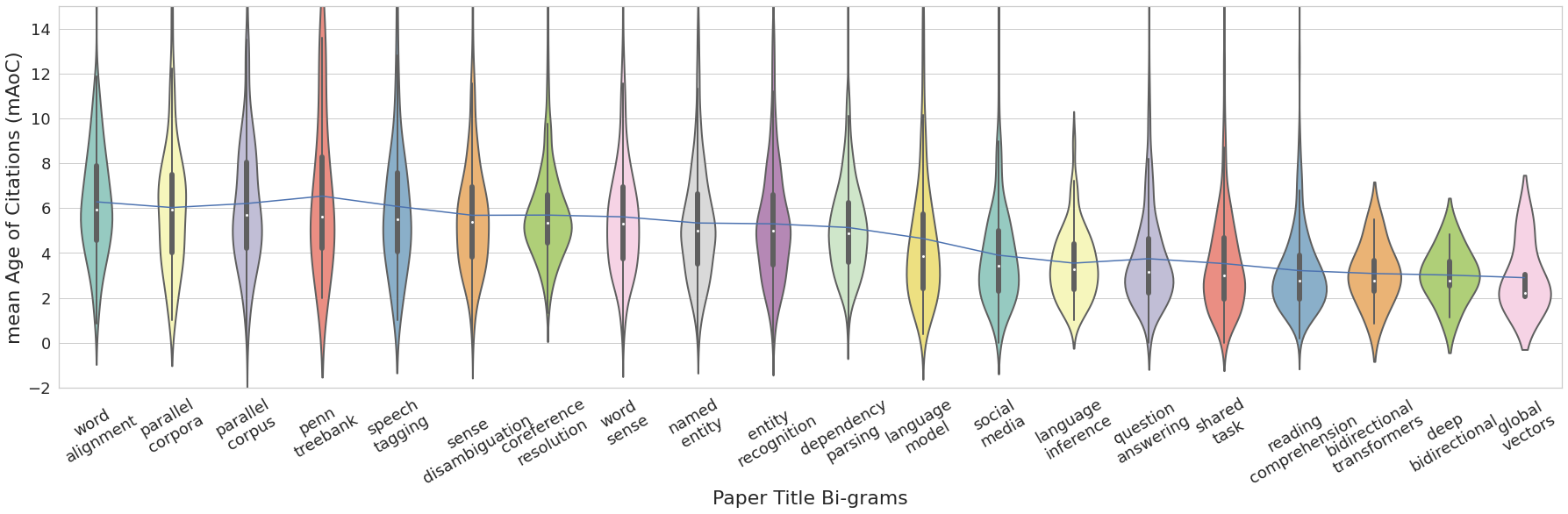}
     \vspace*{-2mm}
    \caption{Distribution of $mAoC$ for frequent bigrams appearing in the titles of citing papers.} 
    % Only selected bigrams are shown here, while distribution for all top-60 frequent bigrams can be found in Figure \ref{fig:r5_new_bigram_distribution_full}.}
    \label{fig:r5_new_bigram_distribution}
    \vspace*{-1mm}
\end{figure*}

\vspace*{6mm}
\noindent
\textbf{Q5. What is the $\pmb{mAoC}$ distribution for different areas within NLP? Relative to each other, which areas tend to cite more older papers and which areas have a strong bias towards recent papers?}\\[-10pt]

\noindent
\textbf{Ans.} 
% In order to study the distribution of $AoC$ across different areas of NLP, we computed the $mAoC$ for all papers in the \textit{AoC dataset}.
The ACL Anthology does not include metadata for sub-areas within NLP. Further, a paper may be associated with more than one area and the distinction between areas can often be fuzzy.
Thus, we follow a rather simple approach used earlier in \citet{mohammad-2020-gender}: using paper title word bigrams as indicators of topics relevant to the paper. 
A paper with \textit{machine translation} in the title is likely to be relevant to the area of machine translation.
Using title bigrams for this analysis also allows for a finer analysis within areas. For example, two bigrams pertaining to finer subareas within the same area can be examined separately. (Papers in different sub-areas of an area need not be similar in terms of the age of the papers they cite.)
% Our approach is similar to the one used by \citet{mohammad-2020-gender} where we considered the top-60 bigrams based on the number of papers sampled for each bigram. 
% Each bigram then loosely represents an area of NLP, allowing us to sample papers from a diverse range of sub-fields within the field. 
% This approach enables us to obtain a comprehensive understanding of the citation patterns across different areas of NLP research.

We first compiled a list of the top 60 most frequent bigrams from the titles of AA papers. 
Next, for each of these bigrams, we created a bin containing all AA papers that had that bigram in their title.\footnote{A single paper may be included in multiple bins.}
For each paper included in any of these bins, we computed $mAoC$. 
Finally, we plotted the distribution of $mAoC$ values for the papers in each bin, as shown in Figure \ref{fig:r5_new_bigram_distribution}. Note that, for the purpose of improving the visibility of the plot, only selected $mAoC$ distributions are depicted in the figure \ref{fig:r5_new_bigram_distribution}.
We then examined the distribution of $mAoC$ for each of these bins.

\paragraph{Results} 
Figure \ref{fig:r5_new_bigram_distribution} shows the $mAoC$ violin plots for each of the bins pertaining to the title bigrams (in decreasing order of median $mAoC$).   
Observe that papers with the title bigrams \textit{word alignment, parallel corpus/corpora, Penn Treebank,  sense disambiguation} and \textit{word sense} (common in the word sense disambiguation area), \textit{speech tagging, coreference resolution, named entity} and \textit{entity recognition} (common in the named entity recognition area), and \textit{dependency parsing} have some of the highest median $mAoC$ (cite more older papers).
In contrast,  papers with the title bigrams \textit{glove vector, BERT pre, deep bidirectional,} and \textit{bidirectional transformers} (which correspond to new technologies)
and papers with title bigrams \textit{reading comprehension, shared task, question answering, language inference, language models,} and \textit{social media} (which correspond to NLP subareas or domains) have some of the lowest median $mAoC$ (cite more recent papers).

% The plots for Gini index and 10-year-old papers cited for these title bigrams are not shown due to space constraints, but will be made available on the project webpage.

% The difference in median $mAoC$, gini coefficient, and percentage of greater than 10-year old papers cited 
% across the considered title bigrams can be as high as [blah, blah] and [blah], respectively.

% From Figure \ref{fig:r5_new_bigram_distribution}, we make several noteworthy observations. 
% Firstly, papers that include deep-learning terminologies, such as BERT, GLoVe, pre-training, neural network, and neural machine, in their title tend to have a lower median $mAoC$ compared to papers that do not include them.
% Moreover, box plots corresponding to such ``areas'' also exhibit a reduced inter-quartile range in comparison to other ``areas''.
% This suggests that papers that utilize these deep-learning techniques tend to cite more recent papers on average.
% Additionally, it is interesting that certain subfields within NLP, such as word-sense disambiguation, speech tagging, coreference resolution, and named entity recognition, demonstrate a higher median $mAoC$ compared to other areas, such as social media analysis, reading comprehension, text classification, and language modeling. 

\paragraph{Discussion} 
The above results suggest that not all NLP subfields are equal in terms of the age of cited papers. In fact, some papers cited markedly more newer papers than others. This could be due to factors such as early adoption or greater applicability of the latest developments, the relative newness of the area itself (possibly enabled by new inventions such as social media), etc.

% have experienced an equal increase in median $mAoC$, and thus the phenomenon recency of citations is not uniform across all subfields.
% The observation from this question prompted an investigation into identifying the prevalent bigrams and unigrams in the cited papers. 
% In the subsequent research question, we delve into this area by exploring the prevalence of bigrams and unigrams in cited papers and examining if any temporal variations exist in these patterns across different year intervals.

% ----------------R5-END---------------

% ----------------R6-START---------------
% Topics/TitleBigrams of cited papers especially pronounced in each period of time.
\vspace*{4mm}
\noindent
\textbf{Q6. What topics are more pronounced in cited papers across different periods of time? }\\[-10pt]

\noindent
\textbf{Ans.} To address this question, we partitioned the research papers % in the \textit{AoC dataset} 
into those published between: 1990--1999, 2000--2009, 2010--2015, and 2016--2021.\footnote{The 2010--2021 period was split into two because of the large number of papers published.} 
% in this period and as it allows for a finer examination.}  
For papers from each period: we first extracted all unigrams and bigrams from the titles of the cited papers. 
Next, for the top 100 most frequent unigrams and bigrams, we calculated the percentage of all citations that had the respective ngram in the cited paper's title --- \textit{the ngram citation percentage}. 
% and were cited in that particular year interval. 

% For the purpose of analysis, we considered the top-x most frequently cited unigrams and bigrams across all intervals.

\paragraph{Results}
% these citation percentages for % the 100 most frequent 

% We also highlight unigrams/bigrams that have undergone substantial changes. 
% A single * indicates that the relative gain from the minimum to the maximum across all year intervals is more than 1500\% for unigrams and 3000\% for bigrams. 
% A double * denotes that the ngram was not cited at all in the interval.
% Note that, for the purpose of improving the visibility, only highlighted ngrams are depicted in the figure \ref{fig:r6_new_bigrams} and \ref{fig:r6_new_unigrams}.

% Our analysis of the bigrams and unigrams highlighted with * in Figure \ref{fig:r6_new_unigrams} and \ref{fig:r6_new_bigrams} revealed several interesting observations. 

% From Figure \ref{fig:r6_new_bigrams} we observe 
Upon examining various bigram citation percentages,
we found that bigrams pertaining to areas such as tree-adjoining grammars have been in decline since the 1990s (cited less as with every subsequent interval). Bigrams pertaining to areas such as conditional random fields and coreference resolution % and error rate 
gained momentum in the middle periods (2000--2016) but have since lost popularity post-2016. 
On the other hand, techniques such as domain adaptation have consistently gained momentum since the 2010s. 
% It is also noteworthy that 
% With the advent of deep learning in NLP, 
Post-2016 keywords related to deep learning technologies such as \textit{convolutional neural nets, deep bi-directional, deep learning, deep neural, Global vectors,} and \textit{jointly learning} experienced a substantial surge in popularity. % since 2016. 
Additionally, certain areas such as cross-lingual and entity recognition %have 
consistently gained momentum since the 1990s.

% We make similar observations from the figure \ref{fig:r6_new_unigrams}. Firstly, 
Upon examining various unigram citation percentages, we found that deep-learning-related terms such as \textit{attention, bert, deep, neural, embeddings,} and \textit{recurrent} saw a substantial increase in citation post-2016. Furthermore, we observed that since the 1990s, there has been a growing trend in NLP papers towards citing research on the social aspects of language processing, as evidenced by the increasing popularity of keywords such as \textit{social} and \textit{sentiment}.

Figures \ref{fig:r6_new_unigrams} and \ref{fig:r6_new_bigrams} in the Appendix show a number of unigrams and bigrams with the most notable changes in 
the ngram citation percentage
% the percentage of all citations that had the respective ngram in the cited paper's title
% percentage 
across the chosen time intervals. 

% \paragraph{Discussion} 

% ----------------R6-END---------------

% ----------------R7-START---------------

\begin{figure*}
     \centering
     \begin{subfigure}[b]{0.48\linewidth}
         \centering
             \includegraphics[width=\linewidth]{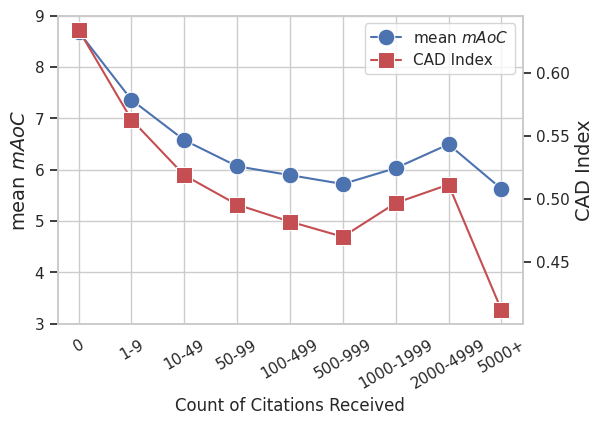}
         \caption{Papers published between 1965 and 2021.}
         
         \label{fig:r7_diversity_overall}
     \end{subfigure}
     \hfill
     \begin{subfigure}[b]{0.48\linewidth}
         \centering
         \includegraphics[width=\linewidth]{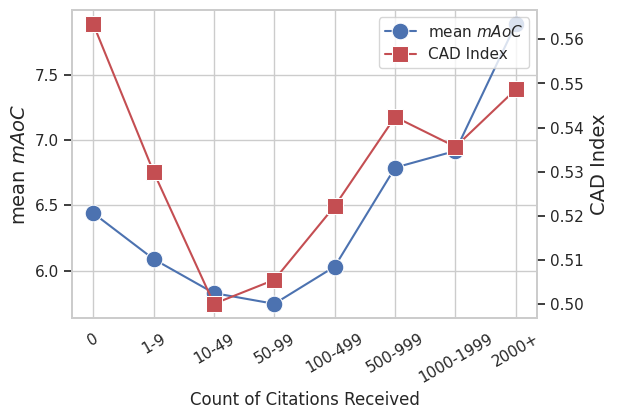}
         \caption{Papers published between 1990 and 2000.}
         \label{fig:r7_diversity_1990}
     \end{subfigure}
     
     \caption{
     %\textbf{ Q7}: 
     Variation of mean $mAoC$ and Citation Age Diversity (CAD) Index (shown on y-axis) for papers with different citation counts (shown on x-axis).}
     \label{fig:r7_diversity}
     \vspace*{-3mm}
\end{figure*}

\vspace{2mm}
\noindent
\textbf{Q7. Do well-cited papers cite more old papers and have more $\pmb{AoC}$ diversity? }

\paragraph{Ans.} We introduce three hypotheses to explore the correlation between temporal citation patterns of target papers and the number of citations the target papers themselves get in the future. %There are several hypotheses:
\\[-18pt]
\begin{enumerate}[label=H\arabic*.]
\item The degree of citation has no correlation with temporal citation patterns of papers.
\vspace*{-2mm}
\item Highly cited papers have more temporal citation diversity than less cited papers.
\vspace*{-2mm}
\item Highly cited papers have less temporal citation diversity than less cited papers.
% \item Degree of citation and temporal citation diversity have a more complex relationship. % (than the options above). 
\vspace*{-2mm}
\end{enumerate}
\noindent Without an empirical experiment, it is difficult to know which hypothesis is true.
H1 seemed likely, however, there were reasons to suspect H2 and H3 also. Perhaps cite more widely is correlated with other factors such the quality of work and thus correlates with higher citations (supporting H2).
Or, perhaps, early work in a new area receives lots of subsequent citations and work in a new area often tends to have limited citation diversity as there is no long history of publications in the area (supporting H3). 

On, Nov 30, 2022, we used the Semantic Scholar API to extract the number of citations for each of the papers in the AoC dataset. 
We divided the AoC papers into nine different bins as per the number of citations: 0, 1--9, 10--49, 50--99, 100--499, 500--999, 1000--1999, 2000--4999, or 5000+ citations. 
For each bin, we calculated the mean of $mAoC$ and \textit{CAD Index}. 
We also computed the Spearman's Rank Correlation between 
% gini-coefficient and 
the \textit{CAD Index} of the citation bins and
the mean of the citation range of each of these bins. % to study the statistical significance.

\paragraph{Results}
Figure \ref{fig:r7_diversity} shows the $mAoC$ and \textit{CAD Index} for each bin (a) for the full AoC dataset, and (b) for the subset of papers published between 1990 and 2000.
% of this analysis are shown in , which presents the mean $mAoC$ and gini-coefficient of $AoC$ for each bin of papers. 
(Figures % \ref{fig:r7_diversity_1990}, 
\ref{fig:r7_diversity_2000} and  \ref{fig:r7_diversity_2010} in the Appendix show plots for papers from two additional time periods.) % the 2000s decade, we created an analogous plot where only papers published in the 1990s, 2000s, and 2010s respectively were considered. 
% Figure \ref{fig:r7_diversity} shows the plot mean $mAoC$ and Gini-coefficient of $AoC$ for each bin of papers. For 
On the full dataset (Figure \ref{fig:r7_diversity_overall}), we observe a clear pattern that the \textit{CAD Index} decreases with increasing citation bin (with the exception of papers in the 1K--2K and 2K--5K bins). 
% Post that there is a decrease in Gini coefficient indicating the increased diversity in the cited papers for bins 1000-1999 and 2000-4999. 
The mean $mAoC$ follows similar trend w.r.t. the \textit{CAD Index}. 

These results show that, for the full dataset, the higher citation count papers tend to have less temporal citation diversity than lower-citation count papers.
However, on the 1990s subset (Figure \ref{fig:r7_diversity_1990}), the \textit{CAD Index} decreased till the citation count < 50 and increased markedly after that. % till 1000 citation count. 
This shows that during the 1990s, the highly cited papers also cited papers more widely in time. Plots for the 2000s and 2010s (Figure \ref{fig:r7_diversity_appendix}) follow a similar trend as the overall plot (Figure \ref{fig:r7_diversity_overall}), indicating that trend of highly cited papers having less temporally diverse citations started around the year 2000.  

\begin{table}[t]
% \small
\begin{center}
\setlength{\tabcolsep}{5pt}%Tighter
% \scalebox{0.95}{
\scalebox{0.8}{
\begin{tabular}{c c c c}
\toprule
 \textbf{1990--99}
& \textbf{2000-09} 
& \textbf{2010--15}
& \textbf{1965--2021 (All)} \\
% & \textbf{Mode} \\
\midrule
0.16  & -1.00$^{*}$ & -0.97$^{*}$ & -0.72$^{*}$\\
% -0.17  & 1.00$^{*}$ & 0.98$^{*}$ & 0.73$^{*}$\\
\bottomrule 
\end{tabular}
}
\end{center}
\vspace*{-3mm}
\caption{\label{tab:q7corr}   
Correlation between the mean number of citations received
% of citation bins 
and \textit{CAD Index} for  
% bins for 
papers from various time periods. The $^{*}$ indicates statistically significant correlation (p-value $<0.05$).}
\vspace*{-5mm}
\end{table}

The Spearman's rank Correlation Coefficients between the mean number of citations for a bin and 
% the Gini coefficient 
the mean $mAoC$ of the citation bins 
% for the bin 
are shown in Table \ref{tab:q7corr}.\footnote{We did not % compute correlations for 
consider 2016--2021 papers because
% the 2016--2021 period because 
% those papers 
they
have had only a few years to accumulate citations.} 
% and reach the larger citation bins.}
% is 0.73 (statistically significant with a p-value $<0.05$). %of 0.0245.
Observe that for the 1990's papers there is essentially no correlation, but there are strong correlations for the 2000s, 2010s, and the full dataset papers. % (a large part of which are papers from 2010 onwards). 

\bl{Similar to Figure \ref{fig:r7_diversity_overall}, in Figure \ref{fig:r7_diversity_area_appendix} (in the Appendix) we show how mean \textit{mAoC} and \textit{CAD Index} of AA papers published between 1965 and 2021 but when broken down by \textit{research topics}. 
This examination across various research topics consistently shows a trend: the higher the citations, the lower the age diversity of citations. This may be because ``mainstream'' work in an area tends to cite lots of other very recent work and brings in proportionately fewer ideas from the past. In contrast, ``non-mainstream'' work tends to incorporate proportionally more ideas from outside, yet receives fewer citations as there may be less future work in that space to cite it. 
% This intriguing trend warrants further exploration and analysis.
}

\paragraph{Discussion}
Papers may receive high citations for a number of reasons; and those that receive high citations are not necessarily model research papers. While they may have some aspects that are appreciated by the community (leading to high citations), they also have flaws. High-citation papers (by definition) are more visible to the broader research community and are likely to influence early researchers more. Thus their strong recency focus in citations is a cause of concern. Multiple anecdotal incidents in the community have suggested how early researchers often consider papers that were published more than two or three years back as "\textit{old papers}". This goes hand-in-hand  with a feeling that they should not cite old papers and therefore, do not need to read them. The lack of temporal citation diversity in recent highly cited papers may be perpetuating such harmful beliefs.

\noindent

% ----------------R7-END---------------

% % ----------------R8-START---------------
% \input{research_questions/RQ8.tex}
% % ----------------R8-END---------------

\section{Demo: CAD Index of Your Paper}
\bl{To encourage authors to be more cognizant of the age of papers they cite, we created an online demonstration page where one can provide the Semantic Scholar ID of any paper and the system returns the number of papers referenced, mean Age of Citation (mAoC), top-5 oldest cited papers, and their years of publication.\footnote{Online demo:  
\anonymousText
{\url{https://huggingface.co/spaces/mrungta8/CitationalAmnesia/}}
{The web link to the online demo is anonymized for review.}
}} Notable, the demo also plots the distribution of mAoC for all the considered papers (all papers published till 2021) and compares it with mean Age of Citation of the input paper. Figure \ref{fig:demo} in the Appendix shows a screenshot of the demo portal for an example input.

\section{Conclusions and Discussion}

% [Summarize things]
% In conclusion, 
This work looks at temporal patterns of citations by presenting a set of comprehensive analyses of the trend in the diversity of age of citations and the percentage of older papers cited in the field of NLP. 
To enable this analysis, we compiled a dataset of papers from the ACL Anthology and their meta-information; notably, the number of citations they received each year since they were published.

We showed that both the diversity of age of citations and the percentage of older papers cited increased from 1990 to 2014, but since then there
 has been a dramatic reversal of the trend. By the year 2021 (the final year of analysis), both the diversity of age of citations and the percentage of older papers cited have reached historical lows. 
We also studied the correlation between the number of citations a paper receives and the diversity of age of cited papers, and found that while there was roughly no correlation in the 1990s, the 2000s marked the beginning of a period where the higher citation levels correlated strongly with lower temporal citation diversity.
% Furthermore, we qualitatively explored the topics that are frequently cited in recent years and found that most of them have not even existed during the 90s, showcasing the rapid transition that the field of NLP is going through. 

% [Discuss Things]
It is a common belief among researchers in the field that the advent of deep neural revolution in the early 2010's has led us to cite more recent papers than before. This analysis confirms and quantifies the extent to which temporal diversity is reduced in this recent period. 
In fact, it shows that the reduction in temporal diversity of citations is so dramatic that it has wiped out steady gains from 1990 to 2014.
While some amount of increased focus on recent papers is expected (and perhaps beneficial) after large technological advances, an open question, now, is whether, as a community, we have gone too far, ignoring important older work.
Our work calls for an urgent need for reflection on the intense recency focus in NLP: How are we contributing to this as researchers, advisors, reviewers, area chairs, and funding agencies?\footnote{This paper cites 16 papers published ten or more years back (35\% of the cited papers).}

\section{Ethics Statement}
This paper analyses scientific literature at an aggregate level. The ACL Anthology freely provides information about NLP papers, such as their title, authors, and year of publication. We do not make use of or redistribute any copyrighted information. All of the analyses
in this work are at aggregate-level, and not about individual papers or authors. In fact, we desist from showing any breakdown of results involving 30 or fewer papers to avoid singling out a small group of papers.

\section{Limitation}
A limitation of this study is that it is based solely on papers published in the ACL Anthology, which primarily represents the international English-language NLP conference community.
While the ACL Anthology is a reputable source of NLP research, it should be acknowledged that a significant amount of research is also published in other venues such as AAAI, ICLR, ICML, and WWW. 
Additionally, there are also vibrant local NLP communities and venues, often publishing in non-English languages, that are not represented in the ACL Anthology. 
As a result, the conclusions drawn from our experiments may not fully capture the global landscape of NLP research and further work is needed to explore the diversity of sub-communities and venues across the world.

This work focuses on the aggregate trends of citing older work in NLP, but does not investigate the reasons for lower citation of certain older papers. 
There may be various factors that contribute to this, such as the accessibility to these older papers, the large number of recent papers, the applicability of these old works, and the technical relevance of the older work. 
Determining the relative impact of each reason is a challenging task. 
Therefore, more research is needed to fully understand the underlying mechanisms that influence the citation of older NLP papers.

% Results of this study 
% % are viewed from a collective perspective and 
% may be affected by various factors such as the author's status, affiliation, geographical location, and many others. 
% While these factors could provide valuable insights, performing the experiments while considering these factors would be complex and challenging to disentangle. 
% Therefore, it should be acknowledged that the results presented in this paper may not fully capture the nuances and complexities of the field of NLP.

This study aims to investigate the factors that contribute to the citation of older works in the field of NLP. We have analyzed different factors such as the mean age of citation, diversity in the age of citations, venue of publication, and subfield of research. Our results indicate that these factors are associated with the citation of older works, but it should be noted that these associations do not establish any causal relationship between them. % Further research is needed to fully understand the underlying mechanisms that influence the citation of older NLP papers.

Lastly, it is important to note that citations can be heterogeneous and can be categorized in different ways. 
For example, some classifications of citations include background, method, and result citations. 
However, certain citations may be more important than others, as shown by previous research such as "\emph{Identifying Meaningful Citations}" by \cite{ValenzuelaEscarcega2015IdentifyingMC}. 
% We acknowledge that we have not distinguished between different types of citations in this work, and urge future work to further examine it. 
% It should be acknowledged that in this current study, we have not distinguished between different types of citations.

\section*{Acknowledgments}
% Anonymized.
\anonymousText
{Many thanks to Roland Kuhn, Rebecca Knowles, and Tara Small for thoughtful discussions.}
{Anonymized.}

\bibliography{anthology,custom}
\bibliographystyle{acl_natbib}

\appendix

% -----------Q3 stufff-----------

\begin{figure*}[!t]
    \centering    \includegraphics[width=2\columnwidth,trim=0 0 0 0,clip]{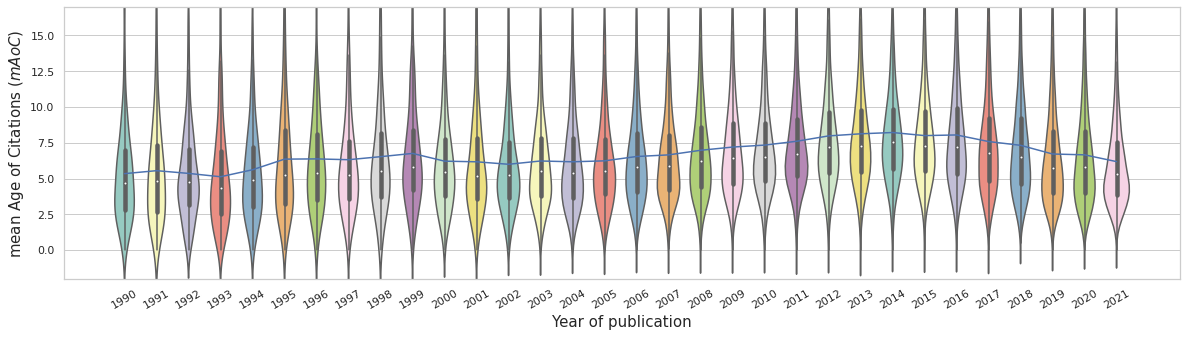} % left bottom right  top | trim order
    \caption{ Distribution of $mAoC$ for papers published between 1990 and 2021.} 
    % A breakdown of these distributions by publication type can be found in Figure \ref{fig:r3_new_aoc_dist_venue}.
    %}
    \label{fig:r3_new_trend_aoc}
\end{figure*}

% \begin{figure*}[t]
%     \centering
%     \includegraphics[width=2\columnwidth]{figure_final/r3_aoc_venue_trend.png} % left bottom right  top | trim order
%     \caption{ Distribution of $mAoC$ for the papers published between 1990 and 2021, broken down by publication type.}
%     \label{fig:r3_new_aoc_dist_venue}
% \end{figure*}
% ------------------------------

\section{Supplementary Statistics and Plots}
\label{sec:supplementary}

In addition to the primary results presented in the main body of the paper, here, we describe included supplementary material in the form of additional statistics and plots.

\subsection{Q3 Results Supplement: Distribution of $\pmb{mAoC}$ Over Years}
\label{sec:supplementary_1}

Figure \ref{fig:r3_new_trend_aoc} shows the violin plots for distributions of $mAoC$ across various years.
% The distribution of $mAoC$ bucketed across different years, as shown by violin plots in Figure \ref{fig:r3_new_trend_aoc}. 
% For example, 
If a paper $x$ was published in year $t$, then $mAoC(x)$ will be a data point for plotting the distribution for year $t$.
The median $mAoC$ for a given year (marked with a white dot within the grey rectangle) reflects the recency of citations, with a lower median $mAoC$ indicating that papers published in that year have cited relatively recent papers. 

The two halves of the grey rectangle on either side of the median correspond to the second and third quartiles. Observe that the third quartile is always longer (spread across more years than the second quartile. This shows that the rate at which papers are cited is higher in years before the median than in the years after the median.
The violin plots indicate that the distributions have a single peak in each of the years considered.

Observe that the median $mAoC$ has an increasing trend from 1990 to 2014 (a trend towards citing more older papers) with the exception of a period between 1998 and 2004 when the median decreased.
However, most notably, from 2014 onward the median $mAoC$ decreased markedly with every year. 
(The median $mAoC$ in 2021 is nearly 2.5 years less than that of 2014.) 

The blue line in Figure \ref{fig:r3_new_trend_aoc} is the mean $mAoC$.
The mean follows a similar trend as the median, with slight variations. 
In particular, it is consistently higher than the median, indicating that the data is skewed to the right, with a few papers having large $mAoC$ that significantly affect the mean.

% Furthermore, we group the publications into three distinct categories - workshops, conferences, and journals.
% We then examine the $mAoC$ distributions for each category across the years.
% Through this approach, we are able to uncover some intriguing insights, as illustrated in Figure \ref{fig:r3_new_aoc_dist_venue}.
% Firstly, journals have a higher median than conferences and workshops for all the years under consideration. 
% Additionally, we see that journals have an offset from zero, indicating that journals cite older papers in general. 
% Secondly, the distributions for conferences and workshops are roughly similar. 
% Before the 2000s, conferences have a higher median than workshops, but after that, the medians for both of them follow a roughly similar pattern. 
% This showcases greater diversification and a tendency to cite older papers in workshops as well. 
% Thirdly, overall, we can see that the median (indicated by the white dot in Figure \ref{fig:r3_new_aoc_dist_venue}) kept increasing until 2014 and decreased considerably since then. 
% This decrease in median indicates a tendency to cite more recent papers, which can be attributed to the fast-paced outlook of this neural era. 
% Lastly, the range of distribution of the average $AoC$ for journals is comparatively less when compared to conferences and workshops.

\subsection{Q6 Results Supplement: Pronounced Topics in the Cited Papers Across Year Intervals}
\label{sec:supplementary_3}
% fillin some text here

\begin{figure}[t]
    \centering   \includegraphics[width=0.95\columnwidth]{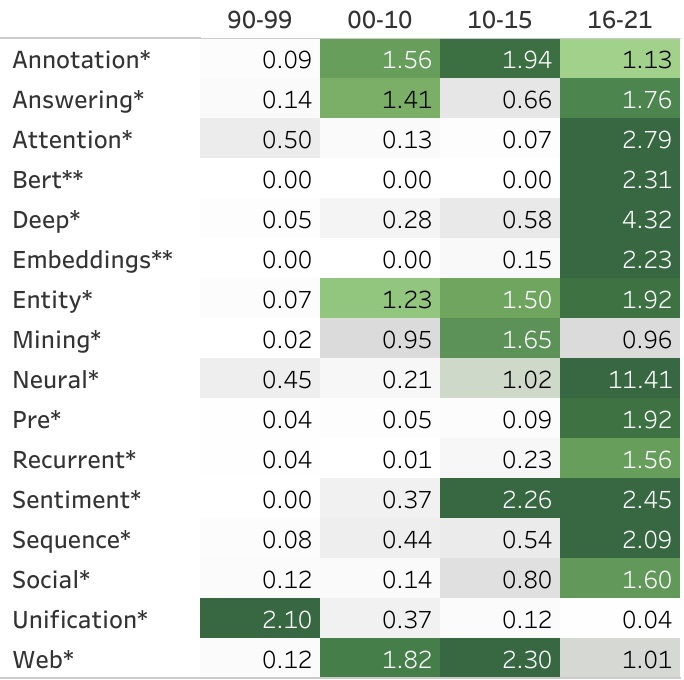}
    \caption{Unigram citation percentages of 
    % the most frequent 
    some notable terms found in the titles of cited papers across different time intervals.
    For example, "Neural"  occurred in 11.41\% of the titles of cited papers in the 2016--2021 interval.
    } 
    % grouped by the publication years of citing paper. Follows the same convention as Figure \ref{fig:r6_new_bigrams}. Note: Among the top-100 overall frequent unigrams, we only show the ones which are highlighted with single or double * as explained in Q6. 90--99: 1990--1999, 00--09: 2000--2009, 10--15: 2010--2015 and 16--21: 2016--2021.}
    \label{fig:r6_new_unigrams}
\end{figure}

\begin{figure}[!ht]
    \centering
    \includegraphics[width=1.01\columnwidth]{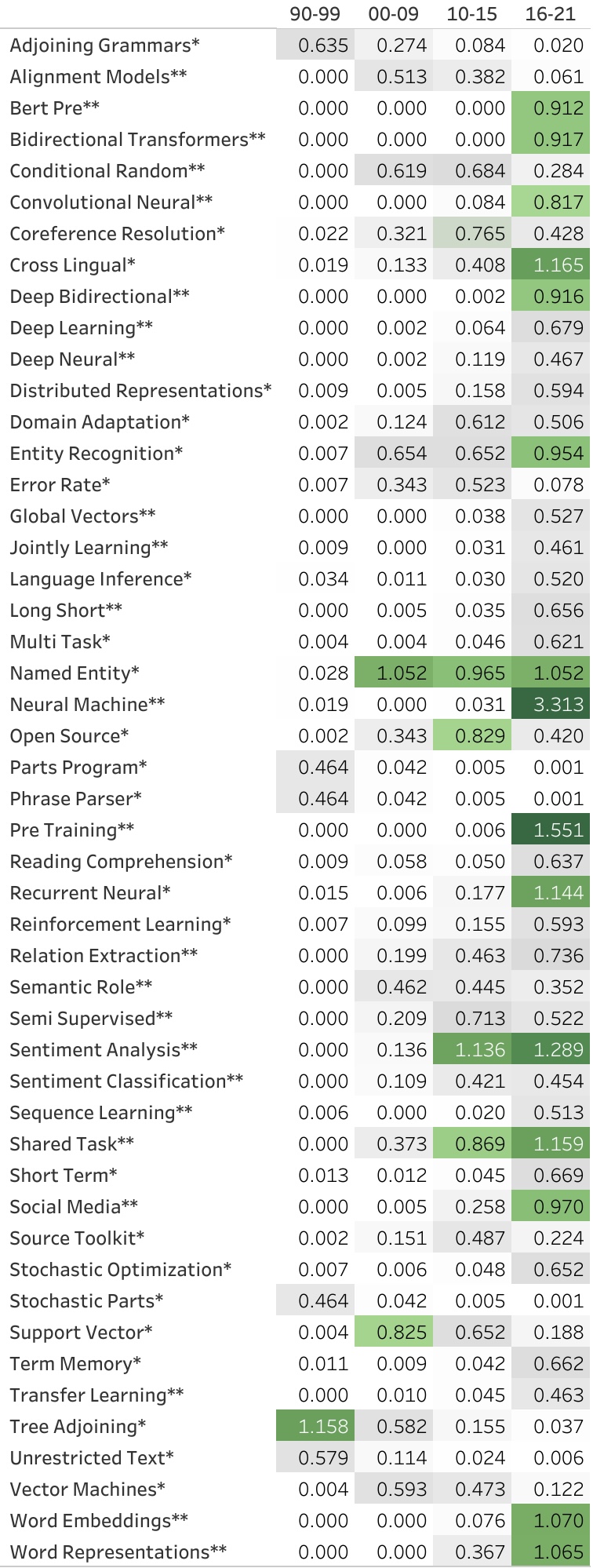}
    \caption{Bigram citation percentages of 
    some notable terms found in the titles of cited papers across different time intervals.
    For example, "Neural Machine"  occurred in 3.313\% of the titles of cited papers in the 2016--2021 interval.
    }
    % the most frequent bigrams found in the title of cited papers, grouped by the publication years of citing paper. 3.313\% of the citations which happened in 2016-2021 had "Neural Machine" in the cited paper's title. Note: Among the top-100 overall frequent bigrams, we only show the ones which are highlighted with single or double * as explained in Q6. 90--99: 1990--1999, 00--09: 2000--2009, 10--15: 2010--2015 and 16--21: 2016--2021.}
    \label{fig:r6_new_bigrams}
\end{figure}

We investigated the distribution of the most frequent unigrams and bigrams (ngrams) found in the title of cited papers, grouped by the publication years of the citing paper. 
% To this end, we computed and illustrated the 
% distribution 
% of these ngrams in 
Figures \ref{fig:r6_new_unigrams} and \ref{fig:r6_new_bigrams}
show the unigrams and bigrams with notable changes in citation percentages across the chosen time intervals. 
% As explained previously in Q6, 
% We highlighted the ngrams which have shown drastic change across different year intervals. 
% In the paper, among the top-100 overall frequent ngrams, we only show the ones which are highlighted with single or double * to better indicate the drastic change. 
A single star (*) indicates that the %relative gain from the minimum to the maximum 
change in the ngram's percentage 
from the minimum interval value to maximum interval value is more than 1500\% for unigrams and 3000\% for bigrams. 
A double star (**) denotes that the ngram was not cited at all in at least one of the intervals.

% As an example of the information presented in these figures, in Figure \ref{fig:r6_new_bigrams}, the cell value corresponding to "Neural Machine" and the interval 2016--2021 denotes that 3.313\% of the citations that occurred in the interval 2016-2021 had "Neural Machine" in the title of the cited paper.

% The extended list and distribution of the data is available on the project's homepage.

\subsection{Q7 Results Supplement: Variation of $\pmb{mAoC}$ and \textit{CAD Index} Across Citation Count Bins}
\label{sec:supplementary_2}

Table \ref{tab:count_citation_paper} shows the number of papers in each citation bin for different segments of papers. We can see that for all the time periods most of the papers have a citation count < 50.

\begin{figure*}
     \centering
     \begin{subfigure}[b]{0.45\linewidth}
         \centering
         \includegraphics[width=\linewidth]{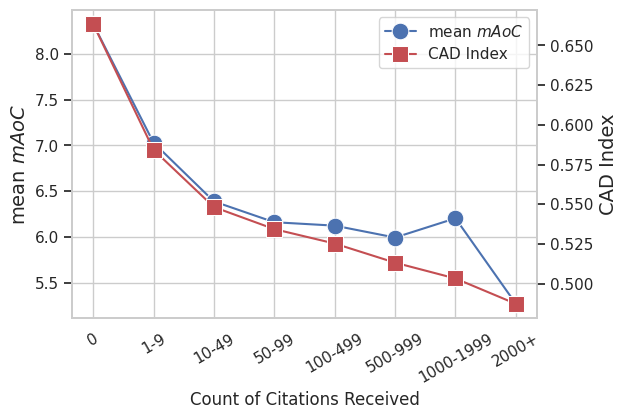}
         \caption{Papers published between 2000 and 2010.}
         \label{fig:r7_diversity_2000}
     \end{subfigure}
     \hfill
     \begin{subfigure}[b]{0.45\linewidth}
         \centering
         \includegraphics[width=\linewidth]{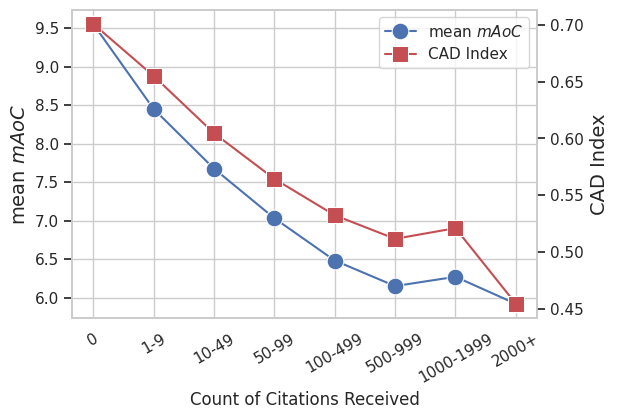}
         \caption{Papers published between 2010 and 2016.}
         \label{fig:r7_diversity_2010}
     \end{subfigure}
     \caption{Variation of mean $mAoC$ and Citation Age Diversity (CAD) (shown on y-axis) for papers with different citation counts (shown on x-axis).}
     \label{fig:r7_diversity_appendix}
\end{figure*}

Figures \ref{fig:r7_diversity_2000} and \ref{fig:r7_diversity_2010}
% We extend the plots discussed in Q7 for the 
show the variation of mean \textit{mAoC} and \textit{CAD Index} for subsets of papers published between 2001 to 2010 and 2011 to 2016, respectively. 
These two plots follow a similar pattern to Figure \ref{fig:r7_diversity_overall} on the full \textit{AoC dataset}. 
The \textit{CAD Index} decreases with increasing the citation bin and the mean $mAoC$ also varies inversely with the citation bin. 

% \bl{Further, similar to Figure \ref{fig:r7_diversity_overall}, in Figure \ref{fig:r7_diversity_area_appendix} we show a  variation of mean \textit{mAoC} and \textit{CAD Index} of AA papers published between 1965 and 2021 but when broken down by \textit{research topics}. 
% This examination across various research topics consistently shows a trend: the higher the citations, the lower the age diversity of citations. This may imply that "mainstream" work in any area tends to cite lots of other very recent work and brings in fewer ideas from the past. In contrast, more "non-mainstream" work tends to incorporate different ideas from outside, yet receives fewer citations as there are less people in that space to cite this kind of work. This intriguing trend warrants further exploration and analysis.}

% -----------Q5 stufff-----------
% \begin{figure*}[!ht]
%     \centering
%     \includegraphics[width=2.1\columnwidth]{figures/r7.png}
%     \caption{Distribution of $mAoC$ for 60 most-frequent bigrams appearing in the title of citing paper.}
%     \label{fig:r5_new_bigram_distribution_full}
% \end{figure*}
% ------------------------------

\begin{table}[t]
% \small
\begin{center}
\setlength{\tabcolsep}{5pt}%Tighter
% \scalebox{0.95}{
\scalebox{0.76}{
% {\small
\begin{tabular}{lrrrr}
\toprule

& \textbf{Full AoC}
& 
& 
&  \\
\textbf{Citation Bin}
& \textbf{1965--2021}
& \textbf{1990--1999} 
& \textbf{2000--09}
& \textbf{2010--15} \\
% & \textbf{Mode} \\
\midrule
\textit{0} &	5559   & 457  & 1062  & 1453 \\
\textit{1--9} &	26794  & 1813  & 5354  & 7090 \\
\textit{10--49} &	 21926  & 1714  & 5804  & 6272 \\
\textit{50--99} &	4843  & 515  & 1517   & 1275 \\
\textit{100--499} &	3860 & 496  &  1296  & 954 \\
\textit{500--999} &	332  & 45   & 105  & 94 \\
\textit{1000--1999} &	123  & 26  &  26  & 49 \\
\textit{2000+} &	106  & 21   & 34   & 27 \\
\bottomrule 
\end{tabular}
}
\end{center}
\vspace*{-3mm}
\caption{\label{tab:count_citation_paper}   
 Number of papers belonging to each citation bin on full AoC dataset, subset of papers published between 1990 to 2000, 2001 to 2010 and 2011 to 2016}
% in AA, also broken down by publication types.}
\vspace*{-3mm}
\end{table}

\begin{figure*}
     \centering
     \begin{subfigure}[b]{0.45\linewidth}
         \centering
         \includegraphics[width=\linewidth]{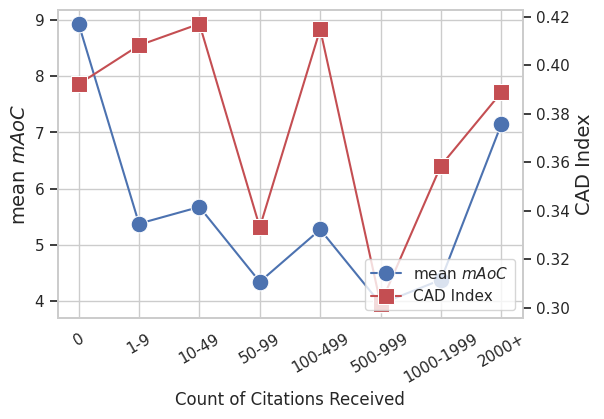}
         \caption{Language inference.}
         \label{fig:r7_diversity_langauge_inference}
     \end{subfigure}
     \hfill
     \begin{subfigure}[b]{0.45\linewidth}
         \centering
         \includegraphics[width=\linewidth]{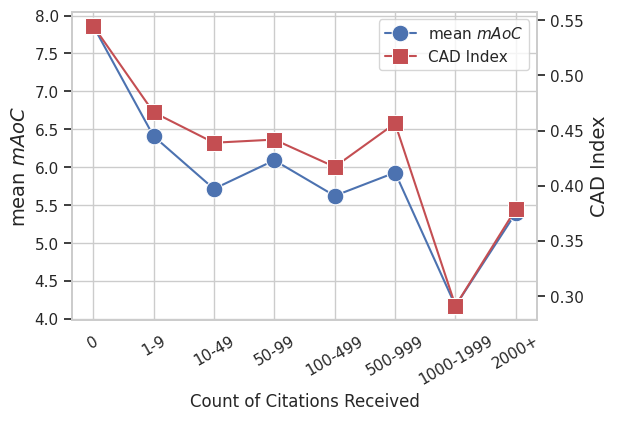}
         \caption{Language model.}
         \label{fig:r7_diversity_langauge_model}
     \end{subfigure}
     
     \begin{subfigure}[b]{0.45\linewidth}
         \centering
         \includegraphics[width=\linewidth]{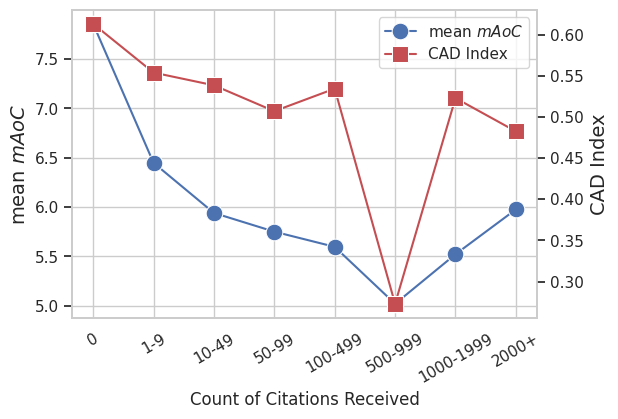}
         \caption{Entity recognition.}
         \label{fig:r7_diversity_entity_recognition}
     \end{subfigure}
    \hfill
     \begin{subfigure}[b]{0.45\linewidth}
         \centering
         \includegraphics[width=\linewidth]{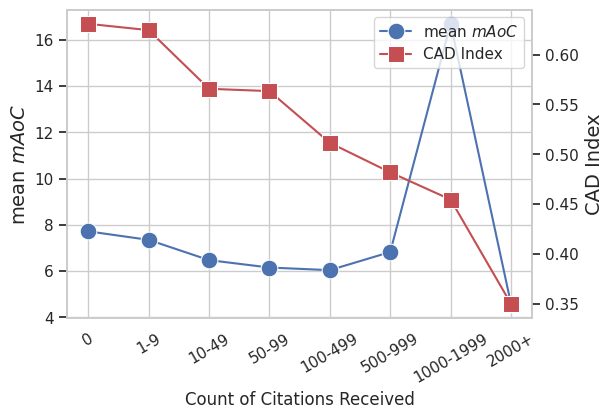}
         \caption{Sense disambiguation.}
         \label{fig:r7_diversity_sense_disambiguation}
     \end{subfigure}

     \begin{subfigure}[b]{0.45\linewidth}
         \centering
         \includegraphics[width=\linewidth]{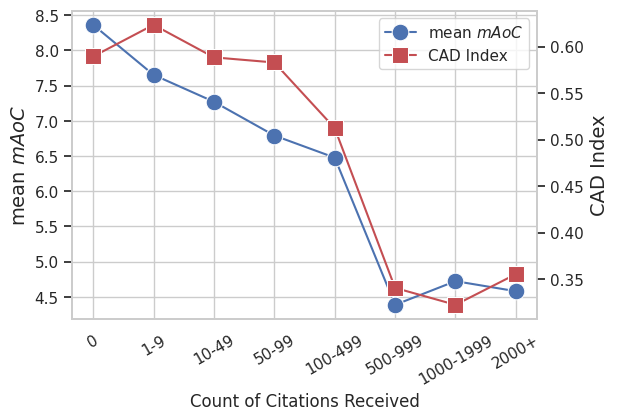}
         \caption{Speech tagging.}
         \label{fig:r7_diversity_speech_tagging}
     \end{subfigure}
     
     \caption{\bl{Variation of mean $mAoC$ and \textit{CAD Index} (shown on y-axis) for papers with different citation counts (shown on x-axis) for papers published between 1965 and 2021 across various \textit{research topics}.}}
     \label{fig:r7_diversity_area_appendix}
\end{figure*}

% \subsection{Demo}
% \bl{We release a working demonstration capturing various metrics proposed in the paper. The demo takes the Semantic Scholar ID of any paper and returns the number of papers referenced, mean Age of Citation (mAoC), top-5 oldest cited paper, and their year of publications. We also plot the distribution of mAoC for all the considered papers (all papers published till 2021) and compare it with the position of the input paper. Figure \ref{fig:demo} shows the overview of the demo portal for an example input.}

% \begin{figure*}[htbp]
%     % \centering    \fbox{\includegraphics[width=2\columnwidth,trim=0 0 0 0,clip]{figure_final/demo.png}} % left bottom right  top | trim order
%     \fbox{\includegraphics[width=1.7\columnwidth,trim=0 2550 135 165,clip]{figure_final/demo-2.pdf}} % left bottom right  top | trim order
%     \caption{Screenshot of the online demo to compute Age-of-Citation metrics for a paper.} 
%     %}
%     \label{fig:demo}
% \end{figure*}

\begin{figure*}[htbp]
    % \floatsetup{heightadjust=object,valign=c}
    \centering
    % \fbox{\includegraphics[width=2\columnwidth,trim=0 0 0 0,clip]{figure_final/demo.png}} % left bottom right  top | trim order
    \fbox{\includegraphics[width=1.7\columnwidth,trim=0 2555 135 168,clip]{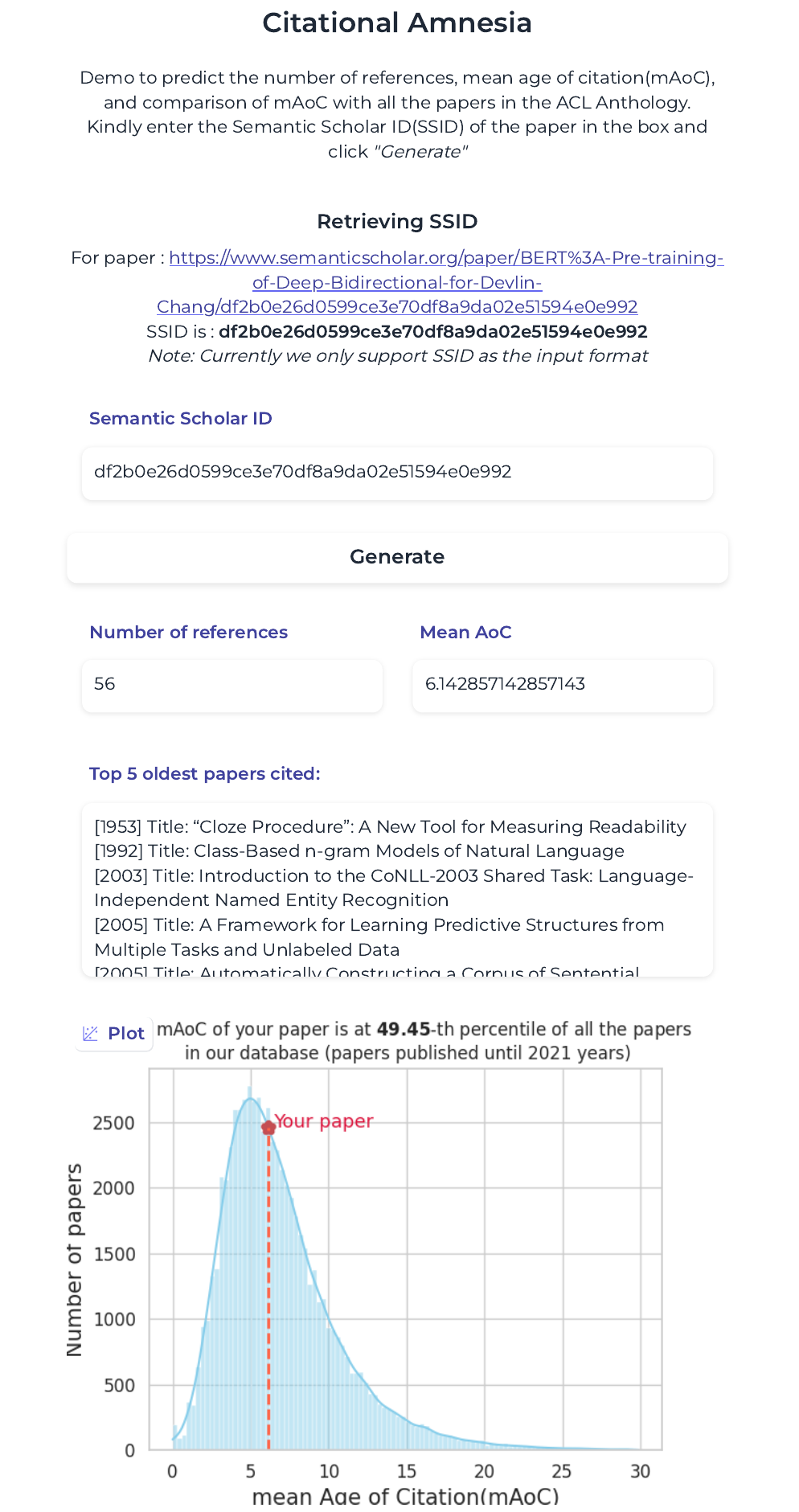
}} % left bottom right  top | trim order
    \caption{Screenshot of the online demo to compute Age-of-Citation metrics for a paper.} 
    \label{fig:demo}
\end{figure*}

% [Create table with 4 rows: Journal Articles, Conference Papers, Workshop Papers, All. The columns can be Mean, Median, Mode. State the year range for the publications that were used to determine the scores.]

\end{document}